\pdfoutput=1
\documentclass[10pt,twocolumn,letterpaper]{article}

\usepackage{cvpr}              
\usepackage{algorithm}
\usepackage{algpseudocode}
\usepackage{footnote}
\usepackage{amsmath}
\usepackage{booktabs}
\usepackage{svg}
\usepackage{amssymb}
\usepackage{bbding}
\usepackage{multirow}
\usepackage{amsmath,amssymb,amsfonts}
\usepackage{graphicx}

\definecolor{cvprblue}{rgb}{0.21,0.49,0.74}
\usepackage[pagebackref,breaklinks,colorlinks,allcolors=cvprblue]{hyperref}

\def\paperID{10809} 
\def\confName{CVPR}
\def\confYear{2025}

\title{Look a Group at Once: Multi-Slide Modeling for Survival Prediction}

\author{
    Xinyang Li$^{1 *}$, Yi Zhang$^{1 *}$, Yi Xie$^{2}$, Jianfei Yang$^{3}$, Xi Wang$^{4}$, Hao Chen$^{5 \dagger}$, Haixian Zhang$^{1 \dagger}$
    \\ Sichuan University$^{1}$, Duke-NUS Medical School$^{2}$, \\ 
    Nanyang Technological University$^{3}$, The Chinese University of Hong Kong$^{4}$, \\  The Hong Kong University of Science and Technology$^{5}$
}

\begin{document}
\maketitle
\renewcommand{\thefootnote}{}
\footnotetext{\hspace{-0.5cm}*These authors contributed equally to this work. \\
$^\dagger$Co-corresponding authors. Hao Chen: \texttt{jhc@ust.hk}; Haixian Zhang: \texttt{zhanghaixian@scu.edu.cn}}

\begin{abstract}
Survival prediction is a critical task in pathology. In clinical practice, pathologists often examine multiple cases, leveraging a broader spectrum of cancer phenotypes to enhance pathological assessment. Despite significant advancements in deep learning, current solutions typically model each slide as a sample, struggling to effectively capture comparable and slide-agnostic pathological features.
In this paper, we introduce GroupMIL, a novel framework inspired by the clinical practice of collective analysis, which models multiple slides as a single sample and organizes groups of patches and slides sequentially to capture cross-slide prognostic features. We also present GPAMamba, a model designed to facilitate intra- and inter-slide feature interactions, effectively capturing local micro-environmental characteristics within slide-level graphs while uncovering essential prognostic patterns across an extended patch sequence within the group framework. Furthermore, we develop a dual-head predictor that delivers comprehensive survival risk and probability assessments for each patient. Extensive empirical evaluations demonstrate that our model significantly outperforms state-of-the-art approaches across five datasets from The Cancer Genome Atlas.
\end{abstract}
\section{Introduction}
\label{sec:intro}
    Survival prediction is a fundamental task in pathology, focused on evaluating patient prognosis through comprehensive analysis of tissue slides and pathological features~\cite{liu2018integrated}. It plays a pivotal role in formulating treatment plans, assessing therapeutic efficacy, and ultimately improving patient outcomes~\cite{fridman2017immune}.
    \begin{figure}[t]
      \centering
      \includegraphics[width=1.0\linewidth]{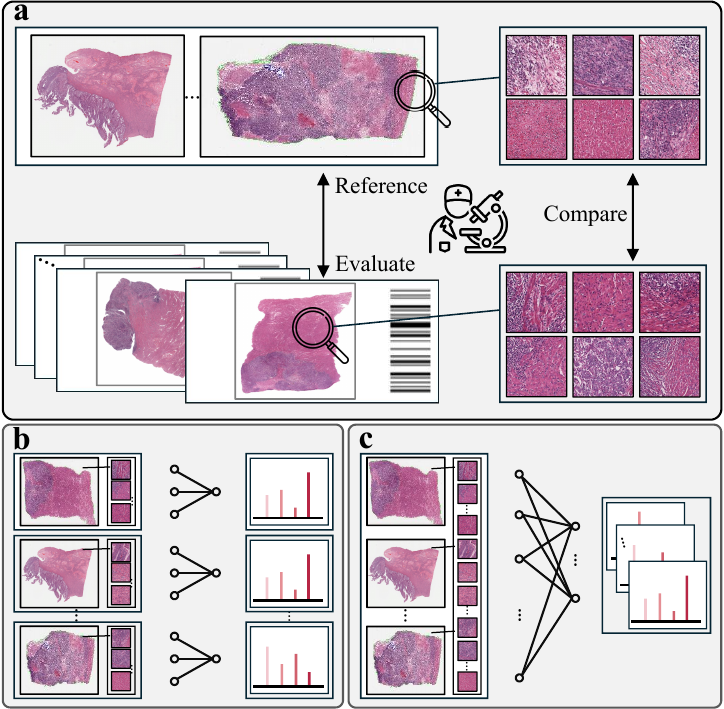}
      \caption{(a) illustrates the process by which pathologists reference other slides and compare phenotypes to enhance their assessments. (b) depicts the conventional procedure of processing WSIs individually. (c) presents our proposed group modeling.}
       \label{fig:teaser}
    \end{figure}
    
    In clinical practice, it is observed that pathologists often benefit from reviewing multiple cases~\cite{kumar2012robbins,zarella2019practical}, as shown in Fig.~\ref{fig:teaser}(a).
    Pathologists suggest that the rationale behind is that different slides reveal a broader range of cancer phenotypes and patient-agnostic prognostic characteristics, which contribute to a non-isolated and comparable prognostic analysis. 
    However, as shown in Fig.~\ref{fig:teaser}(b), existing survival prediction models typically process and predict outcomes for each patient individually, treating each Whole Slide Image (WSI) as a separate sample~\cite{zhu2017wsisa,chen2021whole,shao2023hvtsurv,shao2021weakly,yang2024mambamil,yao2020whole,chen2022scaling,hou2023multi,li2018graph,wang2021hierarchical}. WSIs are segmented into patches, from which features are extracted and aggregated using Multiple Instance Learning (MIL). 
    While these methods learn from individual slides, we assume that the ``collective analysis" can enhance the accuracy and reliability of automatic prognostic evaluation. In this work, as depicted in Fig.~\ref{fig:teaser}(c), we propose to learn from multiple slides at once, rather than from each slide individually, which we refer to as a ``group” of slides.

    To effectively learn from slide groups, two key issues should be addressed: 1) learning from a multitude of patches within a group, and 2) capturing distinctive prognostic features of each slide in the group. To address these issues, we propose: 1) modeling a group of patches as a sequence and introducing the Position and Attention Mamba (PAMamba), which leverages Mamba's~\cite{gu2023mamba} robust ability for handling extremely long sequences and capture distant dependencies, to scan extended patch sequences based on coordinate order and prognostic relevance ranking; and 2) representing each slide as a graph, and developing Graph PAMamba (GPAMamba), which clamps PAMamba by Graph Neural Networks (GNNs), to identify slide-specific characteristics. GPAMamba alternates between aggregating graphs and scanning the sequence, providing a comprehensive understanding of cancer phenotypes and allowing prognostic assessments for each slide to benefit from collective group insights.
    Furthermore, considering the correspondence between survival probability~\cite{zadeh2020bias} and patient risks~\cite{cox1972regression}, we design a dual-head predictor and a combined measure to comprehensively quantify the risk of patients. 

    The main contributions of this work can be summarized as follows:
    \begin{itemize}
        \item Inspired by clinical diagnosis, we propose GroupMIL, the first group-level survival prediction framework to enable cross-slide prognostic analysis.
        \item To capture intra- and inter-slide prognostic features within our group framework, a GNN-empowered two-branch SSM (GPAMamba), is developed.
        \item A dual-head predictor, with an additive loss function, is designed to jointly measure the risk score and survival probability distribution.
        \item Extensive experiments are conducted on five public TCGA datasets, and empirical results show that our model outperforms the state-of-the-art models, demonstrating promise in interpretability, patient stratification, and clinical consistency.
    \end{itemize}

\begin{figure*}[htp]
  \centering
   \includegraphics[width=\linewidth]{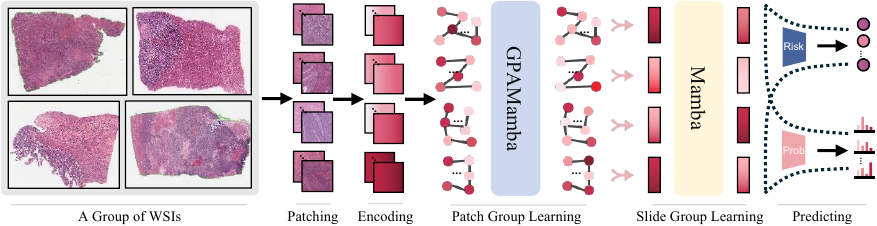}
    \caption{In our framework, each group of slides is segmented into patches, encoded, and represented as independent graphs before being input into GPAMamba. Within GPAMamba, node features are aggregated and sequentially scanned (details in Sec.~\ref{subsec:PAMamba}). The resulting graphs are then pooled to obtain slide representations, which are also arranged sequentially before being analyzed by a Mamba module. Finally, these representations are evaluated using the dual-head predictor, comprising $\mathcal{H}_{\text{risk}}$ and $\mathcal{H}_{\text{prob}}$ (details in Sec.~\ref{subsec:heads}).}
   \label{fig:model}
\end{figure*}
\section{Related Work}
\label{sec:related work}
Survival prediction using WSI was approached as a regression problem~\cite{zhu2016deep}, utilizing pathologist-annotated Regions of Interest (RoIs). This task was later reformulated as a MIL problem, eliminating the need for manual RoI delineation. In these studies, some utilized plain images, while others structurized the images as graphs or sequences.

\vspace{1mm}

\noindent\textbf{Plain Image-based Models.} Convolutional Neural Networks (CNNs) are employed to learn from images. For instance, Zhu~et~al.~\cite{zhu2016deep} and Yao~et~al.~\cite{yao2019deep} sampled patches from WSIs and clustered them based on phenotypes, making final predictions using the fused features of these phenotype clusters. Similarly, Shao~et~al.~\cite{shao2021weakly} employed a grid-based approach to obtain patches, a method that is now widely adopted. Fan~et~al.~\cite{fan2021learning} introduced a pairwise learning strategy that combines features from multiple WSI images of a pair of patients for contrastive learning, enhancing the model's discriminative ability.

\noindent\textbf{Graph-based Models.} Considering the significance of the topological features of WSIs, which plain image-based models tend to overlook, Li~et~al.~\cite{li2018graph}, Chen~et~al.~\cite{chen2021whole}, and Mackenzie~et~al.~\cite{mackenzie2022neural} modeled WSI images as graphs based on the locations of patches and utilized Graph Neural Networks (GNNs) to learn from pre-extracted features. Wang~et~al.~\cite{wang2021hierarchical} constructed multi-scale graphs that span from cells to patches, enabling a hierarchical learning approach for WSI images. Additionally, Wang~et~al.~\cite{wang2023explainable} proposed a coarse-to-fine two-stage graph construction strategy to refine key graph nodes.

\noindent\textbf{Sequence-based Models.} Given the large number of patches, some studies model WSIs as sequences and employ sequence models such as Transformers~\cite{chen2022scaling, fan2022cancer, ding2023pathology, li2023survival, shao2023hvtsurv} and Mamba~\cite{yang2024mambamil} to handle them. Specifically, Chen et al.~\cite{chen2022scaling} adopted a hierarchical fusion strategy of ``cell-patch-region" to learn information from the entire WSI, with each level processed by a vision transformer~\cite{dosovitskiy2020image}. A similar approach was utilized by Shao~et~al.~\cite{shao2023hvtsurv}. Fan~et~al.~\cite{fan2022cancer} expanded their previous work using Transformers~\cite{vaswani2017attention}. Additionally, studies focused on multimodal fusion predictions based on Transformers include \cite{ding2023pathology} and \cite{li2023survival}. Recently, the emergence of Mamba~\cite{gu2023mamba} has pushed the boundaries of sequence models, with Yang et al.~\cite{yang2024mambamil} employing it to achieve improved patch sequence learning.

Graph models are suited for aggregating local information, enhancing the understanding of micro-environmental features, while sequence models capture broader correlations between patches. In this paper, we model WSIs as graphs to capture intra-slide topological information and represent groups as sequences to account for inter-slide global associations, enabling a more comprehensive prognostic analysis.
\section{Method}
\subsection{Preliminaries}
\label{subsec:3.1}
\begin{figure}[H]
    \centering
    \includegraphics[width=1\linewidth]{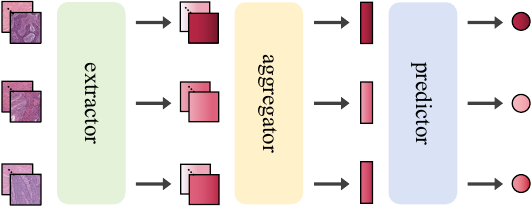}
    \caption{The standard pipeline for survival prediction.}
    \label{fig:pipeline}
\end{figure}

The standard pipeline for survival prediction treats each slide as a sample, processing them in parallel, as illustrated in Fig.~\ref{fig:pipeline}. WSIs are segmented into patches that are encoded into embeddings using a feature extractor. These embeddings are passed through an aggregator to generate slide-level representations, which are subsequently fed into a predictor for the final prediction. The extractor, aggregator, and predictor are described in detail below.

\noindent\textbf{Extractor.} Given a set of WSIs $\{w_1,\dots,w_n\}$, each slide \( w_i\) is segmented into \(N_i\) patches, where \( i \) denotes the index of the \( i \)-th WSI. These patches are encoded into \(N_i\) \(D\)-dimensional embeddings, resulting in each \(w_i\) being represented by a slide feature \(s_i \in \mathbb{R}^{N_i \times D}\). Details on our segmentation and encoding process can be found in Sec.~\ref{sec:4.3}.

\noindent\textbf{Aggregator.} There are two common approaches to process slide features: 1) each \(s_i\) in the set $\{s_1,\dots,s_n\}$ is input into survival models individually, as \(N_i\) varies, with the models processing \(s \in \mathbb{R}^{1 \times N_i \times D}\) one at a time~\cite{chen2021whole,shao2023hvtsurv,hou2023multi,gu2023mamba}. 
2) Alternatively, some studies~\cite{fan2021learning,fan2022cancer} filter some of patches to ensure a consistent number, which may lead to information loss, allowing the model to process a batch of slide features \(s \in \mathbb{R}^{B \times N \times D}\) in parallel, where $B$ represents the batch size and $N$ represents the fixed number of patches per slide. 
Subsequently, a \textit{pooling} operator is used to aggregate \(s\) into slide representations \(s'\) (for vision transformer-based models, these representations are the class tokens), with \(s' \in \mathbb{R}^{1 \times 1 \times D}\) or \(s' \in \mathbb{R}^{B \times 1 \times D}\), where each \(s_i' \in \mathbb{R}^{1 \times D}\). 
Two common practices follow: 1) using $s'$ directly for prediction~\cite{yao2019deep,wang2021hierarchical,yao2020whole}; or 2) passing $s'$ through an additional module for enhancement beforehand~\cite{shao2023hvtsurv,nakhli2023amigo,fan2022cancer}.

In conclusion, prevailing methodologies treat each $w_i$ as a sample for computation, making them unsuitable for holistic prognostic analysis across multiple slides.

\noindent\textbf{Predictor.} 
    The prediction head can be categorized into two types: the risk score head $\mathcal{H}_{\text{risk}}$ and the survival probability head $\mathcal{H}_{\text{prob}}$~\cite{tutz2016modeling,zadeh2020bias}. $\mathcal{H}_{\text{risk}}$ models survival prediction as a regression problem, mapping each $D$-dimensional slide representation \(s'\) to a scalar, the risk score \(r\), which quantifies the patient's risk. The loss function of $\mathcal{H}_{\text{risk}}$, Cox Loss~\cite{cox1972regression}, is defined as:
    \begin{equation}
         \mathcal{L}_{\text{risk}} =  -\sum_{i:E_{i}=1} ( r_i - \log \sum_{j:T_{j}\ge T_{i }} e^{r_{j}} ),   
    \label{eq:Cox Loss}
    \end{equation}
    On the other hand, $\mathcal{H}_{\text{prob}}$ models the task as a classification problem, mapping each \(s'\) to a \(K\)-dimensional vector representing survival probabilities across \(K\) time intervals, predicting the probability of a death event occurring within each interval. The loss function of $\mathcal{H}_{\text{prob}}$, DT Loss~\cite{zadeh2020bias}, is defined as:
    \begin{equation}
    \mathcal{L}_{\text{prob}}= -\sum_{i:E_{i}=1} \log \left( P(T=t_{i} \mid X_{i}) \right) -\sum_{j:E_{j}=0}\log ( S(t_j \mid X_{j})).
    \label{eq:DT Loss}
    \end{equation}
    
    Both heads are widely used. However, $\mathcal{H}_{\text{risk}}$ may show sensitivity to censored patients, while $\mathcal{H}_{\text{prob}}$ discretizes continuous survival time into intervals,  which can potentially lead to information loss and boundary effects.

In this paper, we aim to bring new understandings into the design of both aggregators and predictors, exploring how the group can enhance model performance. First, we introduce the concept of group modeling and describe the organization of two distinct groups in Sec.~\ref{subsec:Group_Modeling}. Next, we detail the design of the aggregator, including the GPAMamba and group fusion module, and how they process their respective groups in Sec.~\ref{subsec:PAMamba}. Finally, we describe our predictor in Sec.~\ref{subsec:heads}. An overview of the proposed framework is provided in Fig.~\ref{fig:model}.

\subsection{Group Modeling}
\label{subsec:Group_Modeling}
Inspired by clinical practice~\cite{kumar2012robbins,zarella2019practical}, we propose the concept of ``group" to enhance cross-slide prognostic analysis. Specifically, given a dataset of WSIs with the same type of cancer, we define every \(B\) random WSIs $\{w_1,\dots,w_b\}$ as a group and treat the entire group, instead of each individual slide, as a single sample. Furthermore, recognizing that pathologists first observe and analyze subtle phenotypes and then compare and draw conclusions about the entire WSI, we model the group hierarchically at both the patch and slide levels.
    
    \noindent\textbf{Patch group.} For a group containing $B$ slide features $ s= \{s_1,\dots,s_b\}$, where each $s_i$ has $N_i$ patches, we sequentialize $s$ to form the patch group \(\ g \in \mathbb{R}^{1 \times \sum_{i=1}^{B}N_{{i}} \times D}\), which represents a single and extended sequence of patches with a length of $\sum_{i=1}^{B}N_{{i}}$. 
    This sequentialization ensures that all patches within the group are visible. Our proposed model, PAMamba, then reorders and rearranges this long patch sequence to uncover slide-agnostic phenotype correlations.

    \noindent\textbf{Slide group.} Similarly, we construct the slide group \(g' \in \mathbb{R}^{1 \times B \times D}\) by sequentializing $\{s'_1, \dots, s'_b\}$, which is also a single sequence of slides of length $B$. Note that $g'$ differs fundamentally in theory from traditional $s' \in \mathbb{R}^{B \times 1 \times D}$ for sequence models. This setting allows for all slides within the group to remain visible. Additionally, we employ another Mamba module to traverse this sequence, thereby enriching $g'$ with cross-slide prognostic information.

By sequentializing the patches and slides hierarchically, we enable collaborative analysis of multiple WSIs, allowing survival models to achieve comprehensive and comparable assessments.

\subsection{Patch Group and Slide Group Learning}
\label{subsec:PAMamba}
\begin{figure}[htp]
  \centering
   \includegraphics[width=1.0\linewidth]{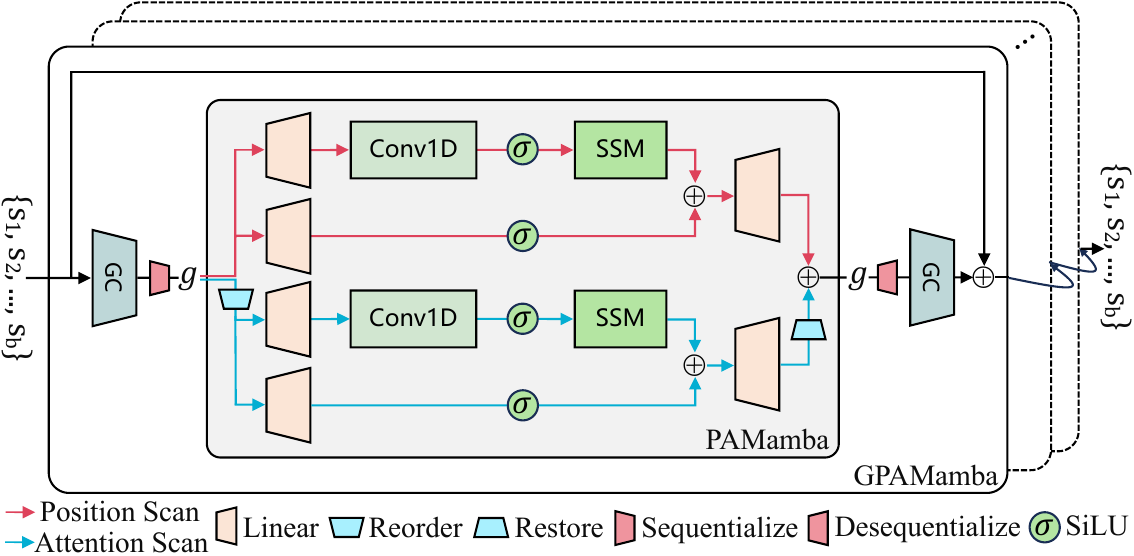}
    \caption{Overview of GPAMamba. GC layers convolve the graphs $\{s_1,...,s_b\}$, while PAMamba scans the sequence $g$. The red lines indicate the position scanning branch, and the blue lines represent the attention scanning branch.}
    \label{fig:Graph PAMamba}
\end{figure}
    
    \noindent\textbf{PAMamba.} The number of patches $N_i$ segmented from a single WSI can reach thousands (at $10\times$ magnification), and that of those within a group can be even greater. 
    Given this, and inspired by Mamba’s efficient handling of long sequences, we design PAMamba to employ position- and attention-based sequential scanning, establishing spatial and prognostic relevance relationships within this extremely long patch sequence. 
    For position sequence $g_{\text{pos}}$, PAMamba scans patches in their spatial order (original order) on the WSI, moving from top to bottom and left to right, connecting WSIs within a group end-to-end. 
    Since groups often contain many patches from normal tissue that may lack strong prognostic significance, PAMamba scans attention sequence $g_{\text{att}}$, which reorders $g$ by the learned weights, to emphasize ``key" patches. 
    Each sequence is processed by a distinct $\phi(x)$, namely $\phi_{\text{pos}}$ and $\phi_{\text{att}}$, respectively, as defined in Eq.~\ref{eq:Mamba_Module}:
    \begin{equation}
   \phi (x)=\mathcal{S}(\sigma (C (L(x))))+ \sigma(L(x)),
    \label{eq:Mamba_Module}
    \end{equation}
    where \(L\) and \(C\) represent a linear layer and a 1D convolution layer, respectively, \(\sigma\) denotes the SiLU activation function~\cite{hendrycks2016gaussian,ramachandran2017searching}, and \(\mathcal{S}\) represents the \textit{SSM} operator~\cite{gu2023mamba}.
    
\begin{algorithm}
\caption{GPAMamba block}
\label{alg:GPAMamba}
\begin{algorithmic}[1]

\Require A group of node features $s=\{s_1, \dots, s_b\}$, where $s_i \in \mathbb{R}^{N_i \times D}$; A group of adjacency matrices $A=\{A_1, \dots, A_b\}$, where $A_i \in \mathbb{B}^{N_i \times N_i}$; Scanning operators $\phi_{\text{pos}} (\cdot)$ and $\phi_{\text{att}}(\cdot)$; GC operators $\mathcal{G}_1(\cdot)$ and $\mathcal{G}_2(\cdot)$; Linear layer $L : \mathbb{R}^D \to \mathbb{R}^1$.

\Ensure Updated node features $s= \{s_1, \dots, s_b\}$.

\For{each graph $i = 1$ to $b$}
    \State $s_i \gets \mathcal{G}_1(A_i, s_i)$ \hfill \Comment{\small Aggregate node features}
    \State $g \gets g.stack(s_i)$ \hfill \Comment{\small Sequentialize $s$}
\EndFor
\State $g_{\text{pos}} \gets \phi_{\text{pos}}(g)$ \hfill \Comment{\small Scan $g$ by $\phi_{\text{pos}}$}
\State $w \gets L(g)$ \hfill \Comment{\small Calculate attention weights}
\State $idx \gets w.argsort()$ \hfill \Comment{\small Get ascending order}
\State $g_{\text{att}} \gets g[idx]$ \hfill \Comment{\small Reorder $g$}
\State $g_{\text{att}} \gets \phi_{\text{att}}(g_{\text{att}})$ \hfill \Comment{\small Scan $g_{\text{att}}$ by $\phi_{\text{att}}$}
\State $g_{\text{att}}[idx] \gets g_{\text{att}}$ \hfill \Comment{\small Restore order}
\State $g \gets (g_{\text{pos}} + g_{\text{att}})$ \hfill \Comment{\small Update $g$}
\For{each graph $i = 1$ to $b$}
    \State $s_i \gets g[:, \sum_{j=1}^{i-1}N_{{j}}:\sum_{j=1}^{i}N_{{j}},:]$ \hfill \Comment{\small Desequentialize $g$}
    \State $s_i \gets \mathcal{G}_2(A_i, s_i)$ \hfill \Comment{\small Update node features}
\EndFor
\State Return $s$
\end{algorithmic}
\end{algorithm}
    
    The process of PAMamba is outlined in the 5 to 11 lines (L) of Alg.~\ref{alg:GPAMamba}. The position sequence \(g_{\text{pos}}\) is handled by the \(\phi_{\text{pos}}\) operator for spatial order learning (L5). 
    Meanwhile, a linear layer generates the attention weights (L6), which reflect the prognostic relevance of the patches (visualization can be found in Sec.~\ref{subsec:visualization}). \(g\) is then reordered in ascending order according to these weights (L7 and L8), and the resulting attention sequence \(g_{\text{att}}\) is processed by the \(\phi_{\text{att}}\) operator for relevant phenotypes learning (L9). Subsequently, to align the corresponding patches between the two sequences, PAMamba restores the processed attention sequence to the original order (L10), and merges the two aligned feature sequences to create a unified and updated group feature $g$ (L11). 
    In summary, PAMamba performs dual-branch long-sequence modeling, enhancing the group-level feature interactions. 
    
    \noindent\textbf{Patch group learning and GPAMamba.} The patch group sequentializes multiple WSIs into a single sample (L3). To preserve the structural integrity and relative independence of each slide, we represent slides as distinct, structurally fixed graphs, where nodes correspond to patches and edges reflect their adjacency relationships (details can be found in Sec.~\ref{sec:4.3}). Each slide corresponds to one graph. Accordingly, Graph Convolutional (GC) layers are used to exchange information between a patch and its neighboring patches within the same slide (L2 and L14), enabling intra-slide micro-environmental learning.
    
    As shown in Fig.~\ref{fig:Graph PAMamba}, the combination of ``GC-PAMamba-GC", along with a desequentialization process to reverse the sequence into graphs (L13), forms the core building block of GPAMamba. This innovative design enables GPAMamba to alternately learn from the $B$ graphs and a single sequence, embodying an ``intra-inter-intra" learning process within the group.
    
    \noindent\textbf{Slide group learning.} The carefully crafted organization and computation procedure of GPAMamba effectively manage patch groups. Here, we employ \(\phi_{\text{int}}\), as defined in Eq.~\ref{eq:Mamba_Module}, to integrate the sequence of group representation $g'$, facilitating the predictor to deliver comparable predictions.

\subsection{Dual-head Predictor and Loss Function}
\label{subsec:heads}
    To harness the rich information within $g'$, we design a dual-head predictor, $\mathcal{H}_{\text{dual}}$, for joint prognostic prediction. As depicted in Fig.~\ref{fig:model}, $\mathcal{H}_{\text{dual}}$ is an intuitive combination of the $\mathcal{H}_{\text{risk}}$ and $\mathcal{H}_{\text{prob}}$ defined in Sec.~\ref{subsec:3.1}.
    Since the cumulative survival probabilities over intervals can serve as a proxy for negative risk, we follow the approach in~\cite{chen2021whole,chen2021multimodal,shao2023hvtsurv}, converting the probability distribution predicted by $\mathcal{H}_{\text{prob}}$ into a risk score. This score is then summed with the predicted risk score $r_i$ output by $\mathcal{H}_{\text{risk}}$, creating a comprehensive measure $R$ that quantifies prognostic outcomes:
    \begin{equation}
    R_i=r_{i} -\sum_{k=1}^{K} P(T > t_k \mid w_i),
    \label{equ:Hazard}
    \end{equation}
    where \(w_i\) denotes a slide in a group, $T$ represents survival time, and $k$ enumerates $K$ time intervals $t$. Thus, \(R\) combines the predicted risk score with probability distributions across each time interval, providing a robust risk assessment with probabilistic detail.
    Accordingly, the dual-head predictor, $\mathcal{H}_{\text{dual}}$, is optimized by a hybrid loss function defined as:
    \begin{equation}
     \mathcal{L}_{\text{overall}}=\alpha \ast  \mathcal{L}_{\text{risk}} + (1-\alpha) \ast  \mathcal{L}_{\text{prob}},
    \label{eq:Loss}
    \end{equation}
where $\mathcal{L}_{\text{risk}}$ and $\mathcal{L}_{\text{prob}}$ are defined in Sec.~\ref{subsec:3.1}.

\section{Experiments}
\begin{table*}
    \centering
    \begin{tabular}{c|cccccc}
    \toprule
    Models & BRCA & BLCA & GBM\&LGG & LUAD & UCEC & Mean\\
    \midrule
    DeepGraphConv~\cite{li2018graph} & 0.577 $\pm$ 0.021 & 0.580 $\pm$ 0.048 & 0.721 $\pm$ 0.031 & 0.580 $\pm$ 0.023 & 0.567 $\pm$ 0.072 & 0.605\\
    DeepAttnMISL~\cite{yao2020whole} & 0.504 $\pm$ 0.042 & 0.524 $\pm$ 0.043 & 0.734 $\pm$ 0.029 & 0.548 $\pm$ 0.050 & 0.597 $\pm$ 0.059 & 0.581\\
    Patch-GCN~\cite{chen2021whole} & 0.598 $\pm$ 0.038 & 0.586 $\pm$ 0.029 & 0.803 $\pm$ 0.021 & 0.545 $\pm$ 0.012 & 0.603 $\pm$ 0.064 & 0.628\\
    ABMIL~\cite{ilse2018attention} & 0.536 $\pm$ 0.038 & 0.564 $\pm$ 0.050 & 0.787 $\pm$ 0.028 & 0.559 $\pm$ 0.060 & 0.625 $\pm$ 0.057 & 0.614\\
    DSMIL~\cite{li2021dual} & 0.603 $\pm$ 0.031 & 0.549 $\pm$ 0.036 & 0.801 $\pm$ 0.023 & 0.594 $\pm$ 0.025 & 0.620 $\pm$ 0.089 & 0.634 \\
    DTFDMIL~\cite{zhang2022dtfd} & 0.643 $\pm$ 0.061 & 0.626 $\pm$ 0.033 & 0.823 $\pm$ 0.038 & 0.644 $\pm$ 0.057 & 0.752 $\pm$ 0.111 & 0.697\\
    TransMIL~\cite{shao2021transmil} & 0.643 $\pm$ 0.027 & 0.610 $\pm$ 0.020 & 0.850 $\pm$ 0.017 & 0.620 $\pm$ 0.027 & 0.711 $\pm$ 0.751 & 0.687\\
    HVTSurv~\cite{shao2023hvtsurv} & 0.624 $\pm$ 0.034 & 0.580 $\pm$ 0.069 & 0.817 $\pm$ 0.021 & 0.620 $\pm$ 0.032 & 0.622 $\pm$ 0.019 & 0.653\\
    S4MIL~\cite{fillioux2023structured} & 0.657 $\pm$ 0.048 & 0.590 $\pm$ 0.059 & 0.850 $\pm$ 0.023 & 0.642 $\pm$ 0.031 & 0.717 $\pm$ 0.109 & 0.691\\
    MambaMIL~\cite{yang2024mambamil} & 0.654 $\pm$ 0.042 & 0.642 $\pm$ 0.038 & 0.861 $\pm$ 0.015 & 0.652 $\pm$ 0.027 & 0.743 $\pm$ 0.055 & 0.710\\
    \midrule
    \textbf{Ours} & \textbf{0.713 $\pm$ 0.052} & \textbf{0.675 $\pm$ 0.018} & \textbf{0.870 $\pm$ 0.020} & \textbf{0.666 $\pm$ 0.033} & \textbf{0.760 $\pm$ 0.047} & \textbf{0.737}\\
    \bottomrule
    \end{tabular}
    \caption{We evaluate the C-index performance of our model against other state-of-the-art models across five datasets. The comparison is conducted with five-fold cross-validation, and the results are reported as the mean and standard deviation across the folds.}
    \label{tab:comparison}
\end{table*}

    \subsection{Datasets}
    We use five publicly available datasets from The Cancer Genome Atlas\footnote{\url{https://portal.gdc.cancer.gov/}} (TCGA)~\cite{kandoth2013mutational}, including pathological slides from different cancers: BReast invasive CArcinoma (BRCA), BLadder urothelial CArcinoma (BLCA), GlioBlastoMa \& Lower Grade Glioma (GBM\&LGG), LUng ADenocarcinoma (LUAD), and Uterine Corpus Endometrial Carcinoma (UCEC).
    
    \subsection{Evaluation Metrics}
    \label{sec:4.2}
    The concordance index (C-index)~\cite{steck2007ranking} is employed as the metric for survival prediction performance of the models. 
    C-index measures the order consistency between event time labels and the predicted outcomes among patients.
    C-index ranges from 0 to 1, where values between 0 and 0.5 indicate poor predictive performance, 0.5 represents complete randomness, and values closer to 1 indicate better predictive performance.
    \subsection{Implementation Details}
    \label{sec:4.3}
    The empirical studies in this work are conducted using the PyTorch 2.0 framework on an NVIDIA GeForce RTX 4090 GPU. We follow CLAM~\cite{lu2021data} for patch segmentation and encoding. Patch segmentation is performed on WSIs at $10\times$ magnification, with each patch sized $512\times512$. 
    These patches are encoded by ResNet50~\cite{he2016deep} pretrained on ImageNet~\cite{deng2009imagenet}, resulting in patch embeddings with a dimensionality \(D\) of 1024. For graph construction, each patch is connected to its 24 nearest neighbors in a $(5 \times 5)$ grid, based on experimental results. During model training and validation, we treat every $6$ slides as a group. We divide the patients' survival time into four intervals.
    We use gradient accumulation with a step size of 32 and the Adam optimizer~\cite{kingma2014adam} with a learning rate of $2\times10^{-4}$. The \(\alpha\) of the loss function $\mathcal{L}_{\text{overall}}$ is set to 0.5.

    \subsection{Comparative Experiments}
    To evaluate our model's performance, we conduct comparative experiments with several SOTA models using the same five-fold cross-validation splits. These include CNN-based models such as ABMIL~\cite{ilse2018attention}, DeepAttnMISL~\cite{yao2020whole}, DSMIL~\cite{li2021dual}, and DTFDMIL~\cite{zhang2022dtfd}; graph-based models like DeepGraphConv~\cite{li2018graph} and Patch-GCN~\cite{chen2021whole}; transformer-based models such as TransMIL~\cite{shao2021transmil} and HVTSurv~\cite{shao2023hvtsurv}; and SSM-based models like S4MIL~\cite{fillioux2023structured} and MambaMIL~\cite{yang2024mambamil}.
    As shown in Tab.~\ref{tab:comparison}, our model outperforms all these models across five datasets.

\subsection{Ablation Studies}
\label{sec: Ablation Studies}

    In this section, we evaluate the effectiveness of each component on the same five-fold splits. 
    The results are presented in Tab.~\ref{tab:Ablation1} and Tab.~\ref{tab:Ablation2}.

\begin{table}[htpb]
    \centering
    \small
    \setlength{\tabcolsep}{2pt}
    \begin{tabular}{cc|cccccc}
    \toprule
    \multicolumn{8}{c}{\textbf{a. Patch Group and Slide Group}} \\
    \midrule
    PG & SG & BRCA & BLCA & G\&L & LUAD & UCEC & Mean \\
    \midrule
    $\boldsymbol{\times}$ & $\boldsymbol{\times}$ & .680$_{\scriptsize .042}$ & .662$_{\scriptsize .021}$ & .855$_{\scriptsize .017}$ & .661$_{\scriptsize .029}$ & .751$_{\scriptsize .055}$ & .722 \\
    \checkmark & $\boldsymbol{\times}$ & .706$_{\scriptsize .057}$ & .665$_{\scriptsize .034}$ & .868$_{\scriptsize .028}$ & .660$_{\scriptsize .036}$ & .750$_{\scriptsize .090}$ & .729 \\
    $\boldsymbol{\times}$ & \checkmark & .689$_{\scriptsize .048}$ & .661$_{\scriptsize .029}$ & .862$_{\scriptsize .018}$ & \textbf{.671}$_{\scriptsize .012}$ & .748$_{\scriptsize .094}$ & .726 \\
    \checkmark & \checkmark & \textbf{.713}$_{\scriptsize .052}$ & \textbf{.675}$_{\scriptsize .018}$ & \textbf{.870}$_{\scriptsize .020}$ & .666$_{\scriptsize .033}$ &  \textbf{.760}$_{\scriptsize .047}$ & \textbf{.737} \\
    \midrule
    \multicolumn{8}{c}{\textbf{b. Graph Representation}} \\
    \midrule
    \multicolumn{2}{c|}{Graph} & BRCA & BLCA & G\&L & LUAD & UCEC & Mean \\
    \midrule
    \multicolumn{2}{c|}{w/o} & \textbf{.721}$_{\scriptsize .075}$ & .667$_{\scriptsize .021}$ & .843$_{\scriptsize .033}$  & .646$_{\scriptsize .035}$  & .706$_{\scriptsize .072}$  & .717\\
    \multicolumn{2}{c|}{\textbf{w/}} & .713$_{\scriptsize .052}$ & \textbf{.675}$_{\scriptsize .018}$ & \textbf{.870}$_{\scriptsize .020}$ & \textbf{.666}$_{\scriptsize .033}$ & \textbf{.760}$_{\scriptsize .047}$ & \textbf{.737} \\
    \midrule
    \multicolumn{8}{c}{\textbf{c. Prediction Head}} \\
    \midrule
    \multicolumn{2}{c|}{Head} & BRCA & BLCA & G\&L & LUAD & UCEC & Mean \\
    \midrule
    \multicolumn{2}{c|}{$\mathcal{H}_{\text{risk}}$} & .705$_{\scriptsize .025}$ & .658$_{\scriptsize .036}$ & .853$_{\scriptsize .017}$ & .662$_{\scriptsize .027}$ & .753$_{\scriptsize .070}$ & .720 \\
    \multicolumn{2}{c|}{$\mathcal{H}_{\text{prob}}$} & .702$_{\scriptsize .077}$ & .666$_{\scriptsize .024}$ & .862$_{\scriptsize .021}$ & .658$_{\scriptsize .022}$ & .713$_{\scriptsize .057}$ & .726 \\
    \multicolumn{2}{c|}{\textbf{$\mathcal{H}_{\text{dual}}$}} & \textbf{.713}$_{\scriptsize .052}$ & \textbf{.675}$_{\scriptsize .018}$ & \textbf{.870}$_{\scriptsize .020}$ & \textbf{.666}$_{\scriptsize .033}$ &  \textbf{.760}$_{\scriptsize .047}$ & \textbf{.737} \\
    \midrule
    \multicolumn{8}{c}{\textbf{d. Group Learning Module}} \\
    \midrule
    \multicolumn{2}{c|}{Module} & BRCA & BLCA & G\&L & LUAD & UCEC & Mean\\
    \midrule
    \multicolumn{2}{c|}{Trans} & .632$_{\scriptsize .043}$  & .663$_{\scriptsize .029}$  & .860$_{\scriptsize .037}$  & .611$_{\scriptsize .032}$  & .710$_{\scriptsize .046}$  & .695 \\
    \multicolumn{2}{c|}{\textbf{Mamba}} & \textbf{.713}$_{\scriptsize .052}$ & \textbf{.675}$_{\scriptsize .018}$ & \textbf{.870}$_{\scriptsize .020}$ & \textbf{.666}$_{\scriptsize .033}$ &  \textbf{.760}$_{\scriptsize .047}$ & \textbf{.737} \\
    \bottomrule 
    \end{tabular}
    \caption{Ablation experiments across five datasets: (a) effect of including or excluding patch and slide groups, with no changes to model structure; (b) comparison of models with and without graph-based WSI representation and GC layers; (c) evaluation of different predictor choices; (d) assessment of the sequence learning module.}
    \label{tab:Ablation1}
\end{table}

    We validate the effectiveness of the proposed Patch Group (PG) and Slide Group (SG). As shown in Tab.~\ref{tab:Ablation1}a, results indicate that PG and SG independently improve mean C-index by 0.8\% and 0.4\%, respectively, with a combined improvement of 1.5\%.
    Additionally, we removed both the graph representations and GC layers, leaving only PAMamba. As shown in Tab.~\ref{tab:Ablation1}b, this led to a 2\% performance decline, underscoring the necessity of graph representation for slides in our group framework.
    Furthermore, we assess the impact of the prediction heads by activating the risk prediction head $\mathcal{H}_{\text{risk}}$ and the probability prediction head $\mathcal{H}_{\text{prob}}$ separately, using original Cox Loss and DT Loss for supervision, respectively. The results in Tab.~\ref{tab:Ablation1}c indicate that the model obtains higher mean C-index with the dual-head predictor $\mathcal{H}_{\text{dual}}$.

\begin{figure}[htpb]
    \centering
    \includegraphics[width=1.0\linewidth]{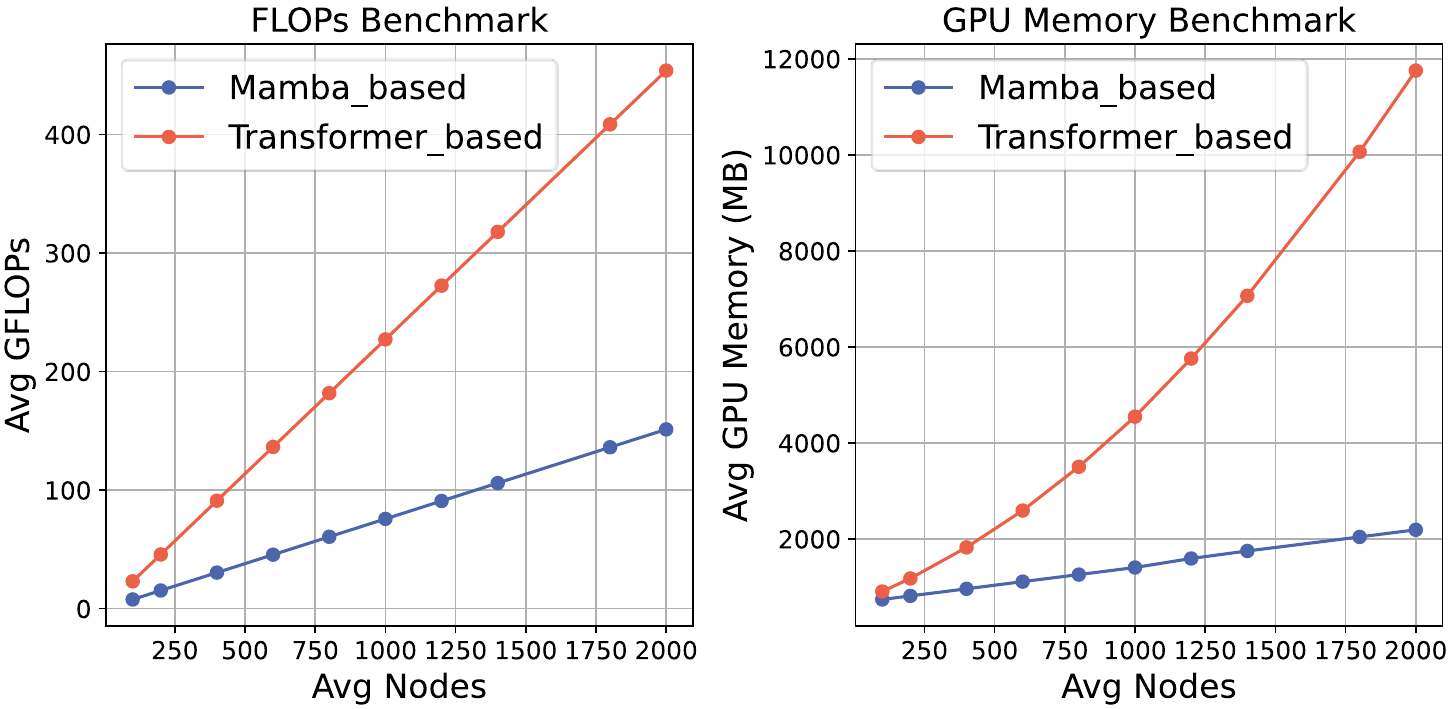}
    \caption{The figure shows a comparison of FLOPs and GPU memory usage between Transformer-based and Mamba-based modules in our framework.}
\label{fig:FLOPS and Memory}
\end{figure}

    We also compared the performance of Mamba with that of Transformer (Trans)~\cite{vaswani2017attention} in learning extended patch sequences. To be specific, we replaced all the three Mamba modules ($\phi_{\text{pos}}$, $\phi_{\text{att}}$, $\phi_{\text{int}}$) with Transformer modules that have a comparable number of parameters. The results, presented in Tab.~\ref{tab:Ablation1}d, demonstrate that the Mamba-based model outperforms the Transformer-based model by 4.2\% overall.
    Additionally, the FLOPs and memory usage are compared under different node counts, with the results available in Fig.~\ref{fig:FLOPS and Memory}. The Mamba-based model demonstrates significant advantages in computational efficiency and resource usage.
    
    We investigate the impact of the two scanning methods in Tab.~\ref{tab:Ablation2}e. The results indicate a 0.5\% difference between the methods individually, while their combination yields an improvement of 1.4\%. Lastly, we show the effect of group size in Tab.~\ref{tab:Ablation2}f. 

\begin{table}[htpb]
    \centering
    \small
    \setlength{\tabcolsep}{2pt}
    \begin{tabular}{c|ccccccc}
    \toprule
    \multicolumn{8}{c}{\textbf{e. Sequence Scanning Method}} \\
    \midrule
    \multicolumn{2}{c|}{Method} & BRCA & BLCA & G\&L & LUAD & UCEC & Mean \\
    \midrule
    \multicolumn{2}{c|}{Pos} & .706$_{\scriptsize .046}$ & .659$_{\scriptsize .022}$ & .861$_{\scriptsize .029}$ & .641$_{\scriptsize .031}$ & .748$_{\scriptsize .055}$ & .723 \\
    \multicolumn{2}{c|}{Att} & .709$_{\scriptsize .038}$ & .663$_{\scriptsize .024}$ & \textbf{.870}$_{\scriptsize .026}$ & .643$_{\scriptsize .045}$ & .753$_{\scriptsize .048}$ & .728 \\
    \multicolumn{2}{c|}{\textbf{Both}} & \textbf{.713}$_{\scriptsize .052}$ & \textbf{.675}$_{\scriptsize .018}$ & \textbf{.870}$_{\scriptsize .020}$ & \textbf{.666}$_{\scriptsize .033}$ & \textbf{.760}$_{\scriptsize .047}$ & \textbf{.737} \\
    \midrule
    \multicolumn{8}{c}{\textbf{f. Group Size}} \\
    \midrule
    \multicolumn{2}{c|}{Size} & BRCA & BLCA & G\&L & LUAD & UCEC & Mean \\
    \midrule
    \multicolumn{2}{c|}{4} & .709$_{\scriptsize .025}$ & .662$_{\scriptsize .031}$ & .865$_{\scriptsize .021}$ & \textbf{.667}$_{\scriptsize .027}$ & \textbf{.764}$_{\scriptsize .049}$ & .733 \\
    \multicolumn{2}{c|}{6} & \textbf{.713}$_{\scriptsize .023}$ & \textbf{.675}$_{\scriptsize .018}$ & .870$_{\scriptsize .020}$ & .666$_{\scriptsize .033}$ & .760$_{\scriptsize .047}$ & \textbf{.737} \\
    \multicolumn{2}{c|}{8} & .704$_{\scriptsize .027}$ & .672$_{\scriptsize .019}$ & \textbf{.872}$_{\scriptsize .035}$ & .654$_{\scriptsize .056}$ & .760$_{\scriptsize .037}$ & .732 \\
    \bottomrule
    \end{tabular}
    \caption{Continuation of Table 2: (e) comparison of sequence scanning methods; (f) impact of group size.}
    \label{tab:Ablation2}
\end{table}

\begin{figure}[t]
  \centering
   \includegraphics[width=\linewidth]{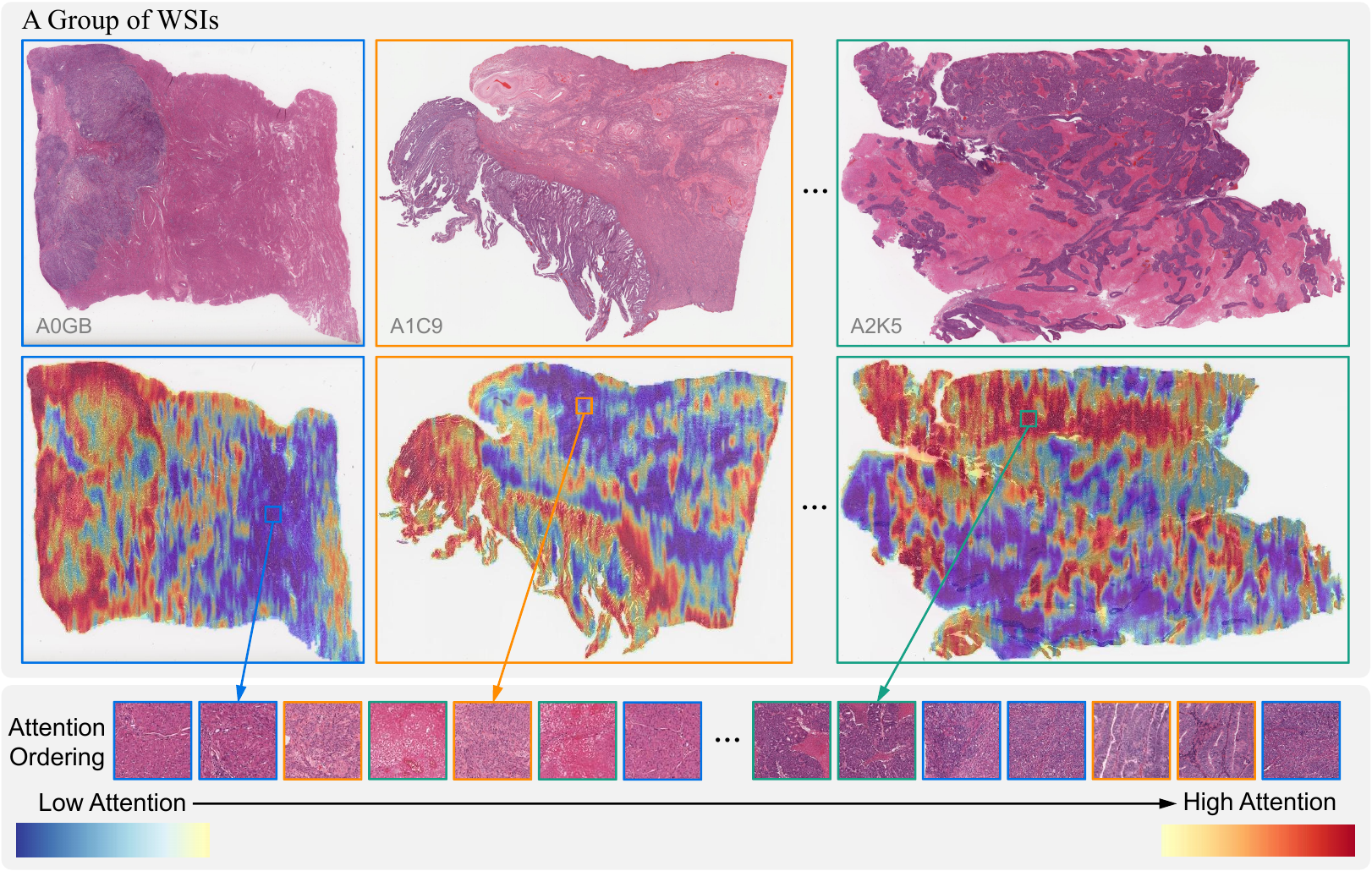}
   \caption{Visualization of a random group from the UCEC dataset. In the heatmaps, red regions indicate areas of high prognostic relevance, while blue regions correspond to lower relevance areas. Patches with the same color border are from the same WSI.}
   \label{fig:Graph_Grad_CAM}
\end{figure}
\begin{figure*}[!h]
    \centering
   \includegraphics[width=\linewidth]{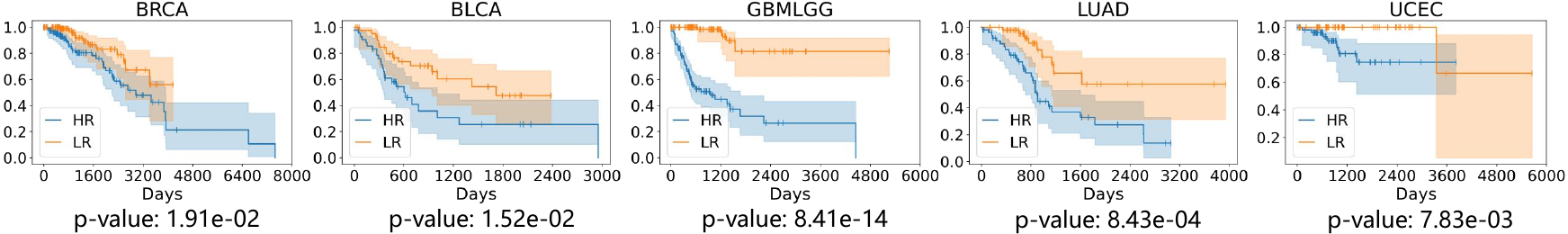}
   \caption{Kaplan-Meier survival curves for high- and low-risk categories (HR and LR) based on the median $R$ across the validation folds of five cancer datasets. Smaller p-values indicate better patient stratification.
}
   \label{fig:KM}
\end{figure*}
\subsection{Group Visualization and Interpretability}
\label{subsec:visualization}
    We visualized the attention sequence \(g_{\text{att}}\) and discussed our findings with pathologists. Fig.~\ref{fig:Graph_Grad_CAM} presents a example from a random group in the UCEC dataset, where the attention distribution of the patches is illustrated through heatmaps. 
    The pathologists noted that the attention weights of \(g_{\text{att}}\) reflect the prognostic relevance of the patches. Patches with low attention are typically derived from normal tissues, such as stroma and muscle, while high-attention patches predominantly originate from tumor and necrotic regions. This distinction highlights the model's capacity to differentiate between benign and malignant tissues, providing visual evidence for survival prediction. 
    
\subsection{Patient Stratification and Clinical Consistency}
    Stratifying patients into different risk categories is essential for targeted treatments and effective patient management. To evaluate our model's capability in this regard, we classify patients into high- and low-risk categories for each cancer type based on the median $R$, which is outlined in Eq.~\ref{equ:Hazard}. We generate Kaplan-Meier (KM) curves~\cite{kaplan1958nonparametric} and conduct log-rank tests~\cite{bland2004logrank}. The results are illustrated in Fig.~\ref{fig:KM}. 
    The p-values across all five datasets are below 0.05, indicating a statistically significant difference between the high- and low-risk cohorts.

\begin{figure*}[t]
  \centering
   \includegraphics[width=\linewidth]{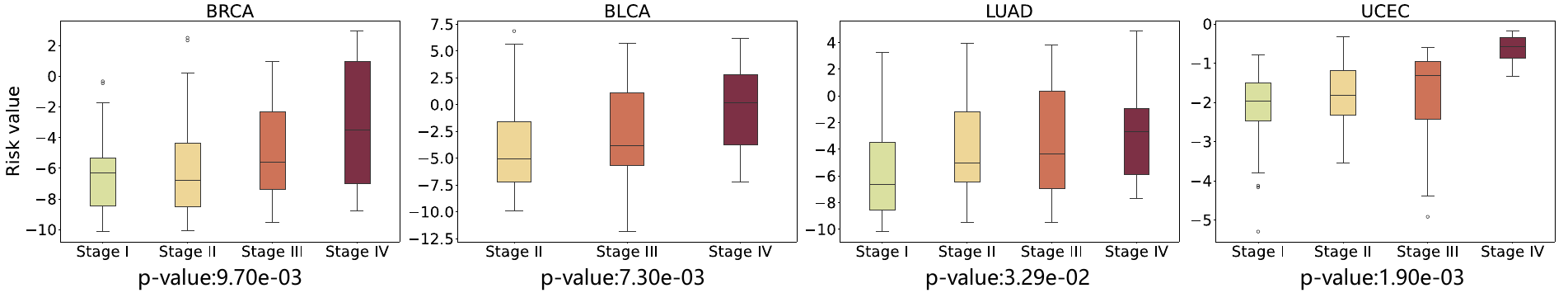}
   \caption{Box plots of the model’s prognostic outcomes, categorized by pathological stages. }
   \label{fig:Staging}
\end{figure*}

    Furthermore, as demonstrated in Fig.~\ref{fig:Staging}, we examine the relationship between patients' pathological stages and their corresponding $R$ values. Due to the unavailability of stage labels for the GBM\&LGG dataset, four plots are presented. The findings reveal a strong correlation between $R$ and clinical pathological staging (p-value $<$ 0.05). This alignment underscores the model's potential to yield reliable prognostic outcomes that align with established clinical assessments.

\section{Discussion}
    \noindent\textbf{How to organize the group?} In our implementation, we randomly group slides of the same cancer type. However, we believe that this grouping method requires further exploration. Organizing slides according to factors such as age, gender, race, region, and genetic information could deepen our understanding of how these variables relate to prognostic predictions, offering valuable insights for cancer diagnosis and treatment.
    
    \begin{table}[h]
    \centering
    \small
    \setlength{\tabcolsep}{1.2pt}
    \begin{tabular}{c|cc|cccccc}
    \toprule
    Model & bs & gs & BRCA & BLCA & G\&L & LUAD & UCEC & Mean \\
    \midrule   
    \multirow{3}{*}{\shortstack{Ours}} & 1 & / & .663$_{\scriptsize .057}$ & \textbf{.646}$_{\scriptsize .021}$ & .862$_{\scriptsize .025}$ & .658$_{\scriptsize .030}$ & .768$_{\scriptsize .094}$ & .719 \\
    & 6 & / & \textbf{.680}$_{\scriptsize .028}$ & .644$_{\scriptsize .028}$ & .858$_{\scriptsize .015}$ & .656$_{\scriptsize .051}$ & .757$_{\scriptsize .064}$ & .719 \\
    \cmidrule{2-9}
    & / & 6 & .676$_{\scriptsize .056}$ & .637$_{\scriptsize .030}$ & \textbf{.868}$_{\scriptsize .020}$ & \textbf{.664}$_{\scriptsize .043}$ & \textbf{.774}$_{\scriptsize .074}$ & \textbf{.724} \\
    \midrule   
    \multirow{3}{*}{\shortstack{Deep-\\Graph-\\Surv\\\cite{li2018graph}}} & 1 & / & .577$_{\scriptsize .021}$ & .580$_{\scriptsize .048}$ & .721$_{\scriptsize .031}$ & \textbf{.580}$_{\scriptsize .023}$ & .567$_{\scriptsize .072}$ & .605 \\
    & 6 & / & \textbf{.600}$_{\scriptsize .021}$ & .569$_{\scriptsize .065}$ & .753$_{\scriptsize .032}$ & .569$_{\scriptsize .043}$ & .526$_{\scriptsize .046}$ & .603 \\
    \cmidrule{2-9}
    & / & 6 & .585$_{\scriptsize .046}$ & \textbf{.582}$_{\scriptsize .057}$ & \textbf{.795}$_{\scriptsize .023}$ & .570$_{\scriptsize .045}$ & \textbf{.596}$_{\scriptsize .052}$ & \textbf{.626} \\
    \midrule
    \multirow{3}{*}{\shortstack{Patch-\\GCN\\\cite{chen2021whole}}} & 1 & / & .598$_{\scriptsize .038}$ & .586$_{\scriptsize .029}$ & .803$_{\scriptsize .021}$ & .548$_{\scriptsize .012}$ & .603$_{\scriptsize .064}$ & .628 \\
    & 6 & / & .588$_{\scriptsize .050}$ & .546$_{\scriptsize .036}$ & \textbf{.807}$_{\scriptsize .021}$ & .543$_{\scriptsize .039}$ & .621$_{\scriptsize .076}$ & .621 \\
    \cmidrule{2-9}
    & / & 6 & \textbf{.667}$_{\scriptsize .031}$ & \textbf{.603}$_{\scriptsize .046}$ & .784$_{\scriptsize .034}$ & \textbf{.603}$_{\scriptsize .015}$ & \textbf{.673}$_{\scriptsize .043}$ & \textbf{.666} \\
    \midrule
    \multirow{3}{*}{\shortstack{Trans-\\MIL\\\cite{shao2021transmil}}} & 1 & / & .643$_{\scriptsize .027}$ & .610$_{\scriptsize .020}$ & \textbf{.850}$_{\scriptsize .017}$ & .620$_{\scriptsize .027}$ & .711$_{\scriptsize .071}$ & .687 \\
    & 6 & / & .664$_{\scriptsize .001}$ & .607$_{\scriptsize .021}$ & .846$_{\scriptsize .016}$ & \textbf{.647}$_{\scriptsize .039}$ & .732$_{\scriptsize .068}$ & .699 \\
    \cmidrule{2-9}
    & / & 6 & \textbf{.665}$_{\scriptsize .022}$ & \textbf{.626}$_{\scriptsize .016}$ & .845$_{\scriptsize .019}$ & .637$_{\scriptsize .035}$ & \textbf{.747}$_{\scriptsize .052}$ & \textbf{.704} \\
    \midrule   
    \multirow{3}{*}{\shortstack{S4-\\MIL\\\cite{fillioux2023structured}}} & 1 & / & \textbf{.657}$_{\scriptsize .048}$ & .590$_{\scriptsize .059}$ & .850$_{\scriptsize .023}$ & \textbf{.642}$_{\scriptsize .031}$ & .717$_{\scriptsize .109}$ & .691 \\
    & 6 & / & .648$_{\scriptsize .013}$ & .583$_{\scriptsize .028}$ & .856$_{\scriptsize .028}$ & .639$_{\scriptsize .028}$ & .715$_{\scriptsize .104}$ & .688 \\
    \cmidrule{2-9}
    & / & 6 & .650$_{\scriptsize .047}$ & \textbf{.591}$_{\scriptsize .025}$ & \textbf{.858}$_{\scriptsize .031}$ & .639$_{\scriptsize .037}$ & \textbf{.739}$_{\scriptsize .038}$ & \textbf{.695} \\
    \midrule
    \multirow{3}{*}{\shortstack{Mamba-\\MIL\\\cite{yang2024mambamil}}} & 1 & / & .654$_{\scriptsize .042}$ & .642$_{\scriptsize .038}$ & .861$_{\scriptsize .015}$ & .652$_{\scriptsize .027}$ & .743$_{\scriptsize .055}$ & .710 \\
    & 6 & / & .656$_{\scriptsize .025}$ & .631$_{\scriptsize .065}$ & .865$_{\scriptsize .037}$ & .657$_{\scriptsize .046}$ & \textbf{.759}$_{\scriptsize .023}$ & .714 \\
    \cmidrule{2-9}
    & / & 6 & \textbf{.666}$_{\scriptsize .035}$ & \textbf{.648}$_{\scriptsize .022}$ & \textbf{.874}$_{\scriptsize .024}$ & \textbf{.660}$_{\scriptsize .033}$ & .752$_{\scriptsize .038}$ & \textbf{.720} \\

    \bottomrule 
    \end{tabular}
    \caption{Model performance under the original settings, batch size (bs) set to 6 and group size (gs) set to 6.}
    \label{tab:universal}
    \end{table}
        
    \noindent\textbf{Does group benefit from a larger batch size?} Due to limitations in the code structure, most survival prediction and computational pathology (CPath) models, as discussed in Sec.~\ref{subsec:3.1}, are restricted to a batch size of only 1. This prompts the question: is the enhanced performance of our proposed methods merely attributable to incorporating more than one sample within the group? To investigate this, we restructured the code for several models to accommodate batch sizes greater than 1, aligning batch size (bs) with group size (gs). Given that these models do not possess an inherent ``slide-level learning module" — such as $\phi_{\text{int}}$ in our framework — that precedes the predictor, we deliberately excluded the slide group here. For a fair comparison, we also omitted $\phi_{\text{int}}$ from our framework; thus, eliminating the entire ``slide group learning" component. All these experiments employ an identical one-head predictor supervised by DT Loss.
    The results, displayed in Tab.~\ref{tab:universal}, do not show a strong correlation between batch size and model performance. 
    Nevertheless, our proposed method achieved the highest mean C-index under identical sizes, underscoring the effectiveness of the group.

    \noindent\textbf{Can group enhance other models?} We evaluated the effectiveness of the group method on several two-stage models, including the graph-based models DeepGraphSurv~\cite{li2018graph} and Patch-GCN~\cite{chen2021whole}, as well as the sequence-based models TransMIL~\cite{shao2021transmil}, S4MIL~\cite{fillioux2023structured}, and MambaMIL~\cite{yang2024mambamil}. The results presented in Tab.~\ref{tab:universal} show that all these models benefit from the group, which may indicate its potential versatility.
    Furthermore, since ``collective analysis" is a common practice in clinical pathological assessment, we believe that group modeling could also benefit related tasks, such as staging and subtyping, in CPath. Relevant work is currently underway, and we hope this modeling approach will inspire researchers and provide pathologists with more comprehensive support across various clinical scenarios.

\section{Conclusion}

    In this paper, we propose a novel group-level survival prediction framework that is in line with clinical observation. Our framework regards a group of slides as a single sample, sequences patches and slides to facilitate inter-slide evaluation. This approach enables each slide to draw on the collective insights of the group, ultimately improving patient outcomes.
    
    To tackle the problems related to extended patch sequences and capturing slide-specific features, we introduce GPAMamba to enhance intra- and inter-slide feature interaction in turn and capture prognostically relevant characteristics. We also develop a dual-head predictor that provides a comprehensive assessment of both risk scores and probability distributions. Extensive experiments carried out on various datasets demonstrate the superiority of our model and the effectiveness of its components. Furthermore, we visualize the WSIs and patches within a group to confirm the model's ability to identify tumor and necrotic regions and their relevance among patients. We also validate the consistency of the model outputs with patient stratification and cancer staging, highlighting its robustness and alignment with clinical assessments.

    In the future, we plan to extend our GroupMIL framework to more CPath tasks. As a general concept to address challenges in CPath and MIL problems, group modeling may offer a solution from new perspectives.
    
{
    \small
    \bibliographystyle{ieeenat_fullname}
    \bibliography{main}
}


\end{document}